\documentclass[11pt,a4paper]{article}
\usepackage[hyperref]{naaclhlt2019}
\usepackage{times}
\usepackage{latexsym}

\usepackage{url}

\usepackage{graphicx}
\usepackage[utf8]{inputenc}
\usepackage{xcolor}
\usepackage{pgfplots} 
\usepackage{booktabs}

\aclfinalcopy 

\title{Measuring Semantic Abstraction of Multilingual NMT \\with Paraphrase Recognition and Generation Tasks}

\author{J{\"o}rg Tiedemann \and  Yves Scherrer \\ Department of Digital Humanities / HELDIG \\ University of Helsinki }

\date{}

\begin{document} 
\maketitle 
\begin{abstract} 
In this paper, we investigate whether multilingual neural translation models learn stronger semantic abstractions of sentences than bilingual ones. We test this hypotheses by measuring the perplexity of such models when applied to paraphrases of the source language. The intuition is that an encoder produces better representations if a decoder is capable of recognizing synonymous sentences in the same language even though the model is never trained for that task. In our setup, we add 16 different auxiliary languages to a bidirectional bilingual baseline model (English-French) and test it with in-domain and out-of-domain paraphrases in English. The results show that the perplexity is significantly reduced in each of the cases, indicating that meaning can be grounded in translation. This is further supported by a study on paraphrase generation that we also include at the end of the paper.
\end{abstract}

\section{Introduction}

An appealing property of encoder-decoder models for machine translation is the effect of compressing information into dense vector-based representations to map source language input onto adequate translations in the target language. However, it is not clear to what extent the model actually needs to model meaning to perform that task; especially for related languages, it is often not necessary to acquire a deep understanding of the input to translate in an adequate way. The intuition that we would like to explore in this paper is based on the assumption that an increasingly difficult training objective will enforce stronger abstractions. In particular, we would like to see whether multilingual machine translation models learn representations that are closer to language-independent meaning representations than bilingual models do. Hence, our hypothesis is that representations learned from multilingual data sets covering a larger linguistic diversity better reflect semantics than representations learned from less diverse material. This hypothesis is supported by the findings of related work focusing on universal sentence representation learning from multilingual data \citep{2018arXiv181210464A,DBLP:journals/corr/abs-1811-01136,Schwenk:2017:repl4nlp} to be used in natural language inference or other downstream tasks.  In contrast to related work, we are not interested in fixed-size sentence representations that can be fed into external classifiers or regression models. Instead, we would like to fully explore the use of the encoded information in the attentive recurrent layers as they are produced by the seq2seq model.

Our basic framework consists of a standard attentional sequence-to-sequence model as commonly used for neural machine translation \citep{DBLP:journals/corr/SennrichFCBHHJL17}, with the multilingual extension proposed by \citet{DBLP:journals/corr/JohnsonSLKWCTVW16}. This extension allows a single system to learn machine translation for several language pairs, and crucially also for language pairs that have not been seen during training.  We use Bible translations for training, in order to keep the genre and content of training data constant across languages, and to enable further studies on increasing levels of linguistic diversity. We propose different setups, all of which share the characteristics of having some source data in English and some target data in English. We can then evaluate these models on their capacity of recognizing and generating English paraphrases, i.e. translating English to English without explicitly learning that task.  Starting with a base model using French--English and English--French training data, we select 16 additional languages as auxiliary information that are added to the base model, each of them separately.

There is a large body of related work on paraphrase generation using machine translation \citep{quirk-etal-emnlp04,finch-etal-jnlp04,prakash-etal-coling16} based on parallel monolingual corpora \citep{mscoco,fader-etal-acl13}, pivot-based translation \citep{bannard-etal-acl05,mallinson-etal-eacl17} and paraphrase databased extracted from parallel corpora \citep{ganitkevitch-etal-naacl13}.  Related work on multilingual sentence representation \citep{2018arXiv181210464A,Schwenk:2017:repl4nlp,DBLP:journals/corr/abs-1901-07291} has focused on fixed-size vector representations that can be used in natural language inference \citep{DBLP:journals/corr/abs-1809-05053,DBLP:journals/corr/abs-1809-04686} or other downstream tasks such as bitext mining \citep{DBLP:journals/corr/abs-1811-01136} or (cross-lingual) document classification \citep{Schwenk:2018:lrec_mldoc}.

\section{Experimental Setup}

For our experiments, we apply a standard attentional sequence-to-sequence model with BPE-based segmentation. We use the Nematus-style models \citep{DBLP:journals/corr/SennrichFCBHHJL17} as implemented in MarianNMT \citep{mariannmt}. These models apply gated recurrent units (GRUs) in the encoder and decoder with a bi-directional RNN on the encoder side. The word embeddings have a dimensionality of 512 and the RNN dimensionality is set to 1,024. We enable layer normalization and we use one RNN layer in both, encoder and decoder.

In training we use dynamic mini-batches to automatically fit the allocated memory (3GB in our case) based on sentence length in the selected sample of data. The optimization procedure applies Adam \citep{Kingma2014adam} with mean cross-entropy as the optimization criterion. We also enable length normalization, exponential smoothing, scaling dropout for the RNN layers with ratio 0.2 and also apply source and target word dropout with ratio 0.1. All of these values are recommended settings that have empirically been found in the related literature. For testing convergence, we use independent development data of roughly 1,000 test examples and BLEU scores to determine the stopping criterion, which is set to five subsequent failures of improving the validation score. The translations are done with a beam search decoder of size 12. The validation frequency is set to run each 2,500 mini-batches.

For the multilingual setup, we follow \citet{DBLP:journals/corr/JohnsonSLKWCTVW16} by adding target language flags to the source text placing them as pseudo tokens in the beginning of each input sentence. We always train models in both directions enabling the model to read and generate the same language without explicitly training that task (i.e. paraphrasing is modeled as zero-shot translation).  BPE \citep{Sennrich2016subword} is used to avoid unknown words and to improve generalisations. Note that in our setup we need to ensure that subword-level segmentations are consistent for each language involved in several translation tasks. We opted for language-dependent BPE models with 10,000 merge operations for each code table. The total vocabulary size then depends on the combination of languages that we use in training but the vocabulary stays exactly the same for each language involved in all experiments.

\begin{table}[t!]
\small
\begin{center}
\begin{tabular}{|l|rrr|}
\hline 
\bf Language & \bf Transl. & \bf Verses & \bf Tokens \\ 
\hline
English    & 19 &  234,173 & 6,750,869\\
French     & 14 &  369,910 & 10,529,929\\
\hline
Afrikaans  & 5  &   75,974 & 2,329,773\\
Albanian   & 2  &   58,192 & 1,648,242\\
Breton     & 1  &    1,781 & 44,316\\
German     & 24 &  499,844 & 13,712,459\\
Greek      & 7  &   87,218 & 2,357,095\\
Frisian    & 1  &   29,173 & 852,582\\
Hindi      & 4  &   93,242 & 2,829,274\\
Italian    & 5  &  122,363 & 3,429,182\\
Dutch      & 3  &   87,460 & 2,596,298\\
Ossetian   & 2  &   37,807 & 936,533\\
Polish     & 5  &   52,668 & 1,248,108\\
Russian    & 5  &   75,904 & 1,727,536\\
Slovene    & 1  &   29,088 & 748,367\\
Spanish    & 8  &  236,830 & 6,607,932\\
Serbian    & 2  &   35,019 & 844,299\\
Swedish    & 1  &   29,088 & 833,983\\
\hline
\end{tabular}
\end{center}
\caption{\label{tab:data} Statistics about the Bible data in our collection: number of individual Bible translations, number of verses and number of tokens per language in the training data sets.} 
\end{table}

\begin{figure*}[t]
\centering
\begin{tikzpicture}
\begin{axis}[
	ybar,
    ymin=0, ymax=60, 
    width=0.53\linewidth, height=0.36\linewidth,
    symbolic x coords={Eng--Fra, +Breton, +Slovene, +Ossetian, +Serbian, +Spanish, +Greek, +Hindi, +Frisian, +Swedish, +Afrikaans, +Polish, +Italian, +Russian, +German, +Dutch, +Albanian
    },
    xtick=data,
    xticklabel style={inner sep=0pt, anchor=north east, rotate=60, font=\small},
    log ticks with fixed point,
    yticklabel style={font=\small}
    ]
   \addplot[ybar legend,fill=black!10] coordinates {
   (Eng--Fra, 55.0338)
   (+Afrikaans, 7.90764)
   (+Albanian, 7.59045)
   (+Breton, 22.4147)
   (+German, 7.6483)
   (+Greek, 8.48242)
   (+Frisian, 8.15503)
   (+Hindi, 8.27394)
   (+Italian, 7.77317)
   (+Dutch, 7.60353)
   (+Ossetian, 8.78013)
   (+Polish, 7.80782)
   (+Russian, 7.73866)
   (+Slovene, 9.9928)
   (+Spanish, 8.66297)
   (+Serbian, 8.66536)
   (+Swedish, 7.94782)
      };
    \draw [red] ({rel axis cs:0,0}|-{axis cs:+Afrikaans,6.05361}) -- ({rel axis cs:1,0}|-{axis cs:+Swedish,6.05361}); \end{axis}
\end{tikzpicture}%
\hfill
\begin{tikzpicture}
\begin{axis}[
	ybar,
    ymin=20, ymax=100,
    width=0.53\linewidth, height=0.36\linewidth,
    symbolic x coords={Eng--Fra, +Breton, +Slovene, +Ossetian, +Serbian, +Spanish, +Greek, +Hindi, +Frisian, +Swedish, +Afrikaans, +Polish, +Italian, +Russian, +German, +Dutch, +Albanian
    },
    xtick=data,
    xticklabel style={inner sep=0pt, anchor=north east, rotate=60, font=\small},
    log ticks with fixed point,
    yticklabel style={font=\small}
    ]
   \addplot[ybar legend,fill=black!10] coordinates {
   (Eng--Fra, 96.1473)
   (+Afrikaans, 42.4788)
   (+Albanian, 43.8201)
   (+Breton, 68.1274)
   (+German, 41.5083)
   (+Greek, 42.5382)
   (+Frisian, 45.3267)
   (+Hindi, 44.8898)
   (+Italian, 40.0764)
   (+Dutch, 42.4013)
   (+Ossetian, 48.8973)
   (+Polish, 48.5963)
   (+Russian, 42.4576)
   (+Slovene, 47.4599)
   (+Spanish, 41.3748)
   (+Serbian, 44.1786)
   (+Swedish, 42.3524)
      };
    \draw [red] ({rel axis cs:0,0}|-{axis cs:+Afrikaans,22.295}) -- ({rel axis cs:1,0}|-{axis cs:+Swedish,22.295});
\end{axis}
\end{tikzpicture}
\caption{\label{fig:ppl} Paraphrase perplexity measured on Bible (left) and Tatoeba (right) test sentences (lower values are better). The figures show the effect of one auxiliary language added to the bilingual French-English model (leftmost bars). The lower red line represents the supervised model trained on English paraphrases. Languages are sorted by decreasing perplexity on the Bible data.}
\end{figure*}
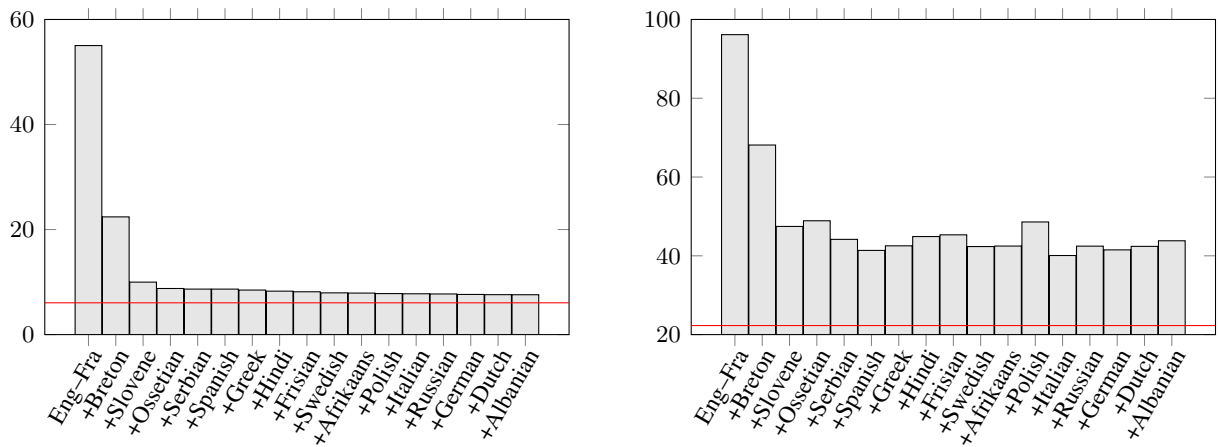

\subsection{Training data and configurations}

The main data we use for our experiments comes from a collection of Bible translations \citep{MAYER14.220.L14-1215} that includes over a thousand languages. For high-density languages like English and French, various alternatives are available (see Table~\ref{tab:data}). Using the Bible makes it possible to easily extend our work with additional languages representing a wide range of linguistic variation, while at the same time keeping genre and content constant across languages.

For the sake of discussion, we selected English as our pivot language that we will use for evaluating the ability of the model to act as a paraphrase model. Furthermore, we took French as a second language to create a bilingual baseline model that can translate in both directions. As additional auxiliary languages, we then apply the ones listed in Table~\ref{tab:data} together with some basic statistics of the training data.  The idea behind the language selection is to create a somewhat diverse set of languages representing different amounts of coverage and typological relationships. The set is easy to extend but training requires extensive resources, which necessarily limits our selection at this point.

In the general setup, we do not include any pairs of English Bible translations as we do not want to evaluate a model that is specifically trained for a paraphrasing task. However, for comparison we also create a model comprising all pairs of English translation variants, which will serve as an upper bound (or rather, a lower bound in terms of perplexity) for models that are trained without explicit paraphrase data.

Exhaustively looking at all possible subsets of languages is not possible even with our small selection of 18 languages. Therefore, we restricted our study to the following test cases:

\noindent\textbf{Bilingual model:} A model trained on all combinations of
  English and French Bible translations. Each pair of aligned Bible
  verses represents two training instances, one for English-to-French
  and one for French-to-English. We also include French-to-French
  training instances using identical sentences in the input and
  output, in order to guide the model to correctly learn the semantics
  of the language flags.\footnote{ \label{fn:frenchfrench} During our
    initial experiments, we realized that the language labels did not
    always pick up the information about the target language they are
    supposed to indicate. Especially in the bilingual case this makes
    sense as the model always sees the same language pair and
    identifying the source language is enough to determine what kind
    of output language it needs to generate. The label is not
    necessary and, therefore, ignored.}

\noindent\textbf{Trilingual models:} Translation models trained on all bilingual
  combinations of Bibles in three languages -- English, French and
  another auxiliary language (in both directions) + identical French
  verse pairs.

\noindent\textbf{Multilingual model:} One model that includes all languages in
  our test set with training data in both directions (translating from
  and to English or French) + identical French verse pairs.

\noindent\textbf{Paraphrase model:} A model trained on combinations of English
  Bible translations (the supervised upper bound).

  Note that all models (including the bilingual one) cover the same English data including all Bible variants. We use exactly the same vocabulary for the English portion of each setup and no new English data is added at any point and any change that we observe when testing with English paraphrase tasks is due to the auxiliary languages that we add to the model as a translational training objective.

\begin{figure*}[t]
\hspace*{-12pt}
\includegraphics[width=0.36\textwidth]{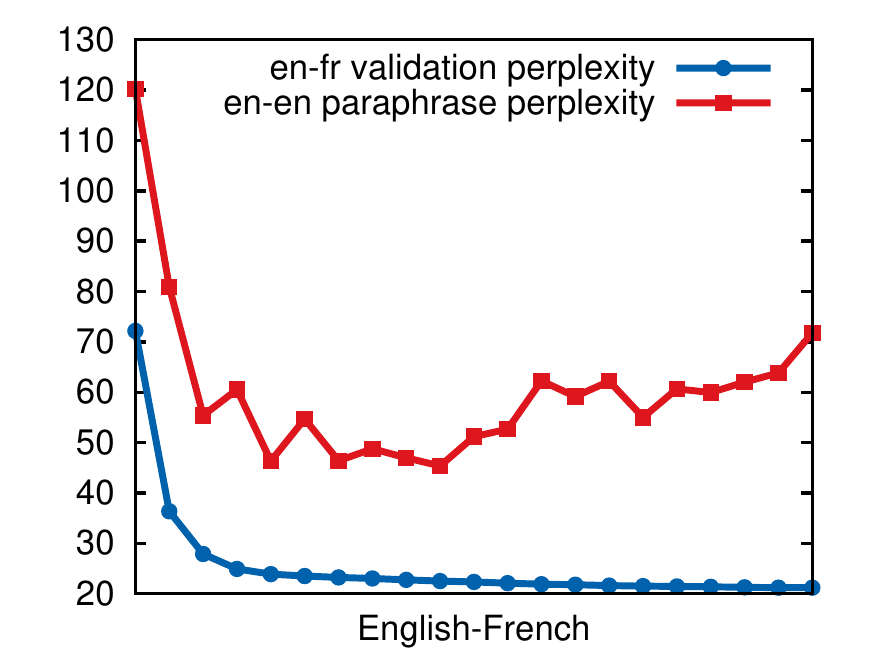}\hspace*{-14pt}
\includegraphics[width=0.36\textwidth]{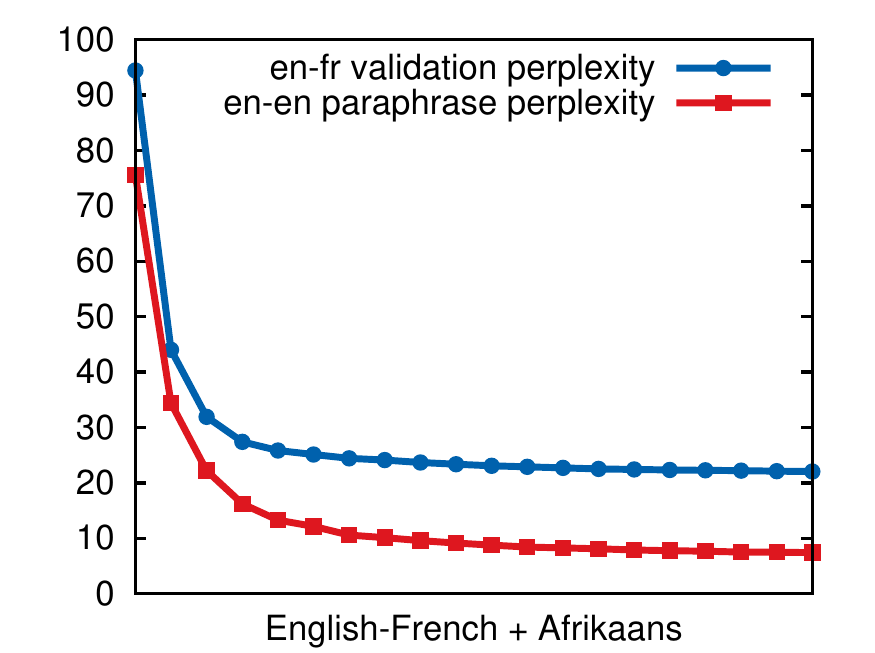}\hspace*{-14pt}
\includegraphics[width=0.36\textwidth]{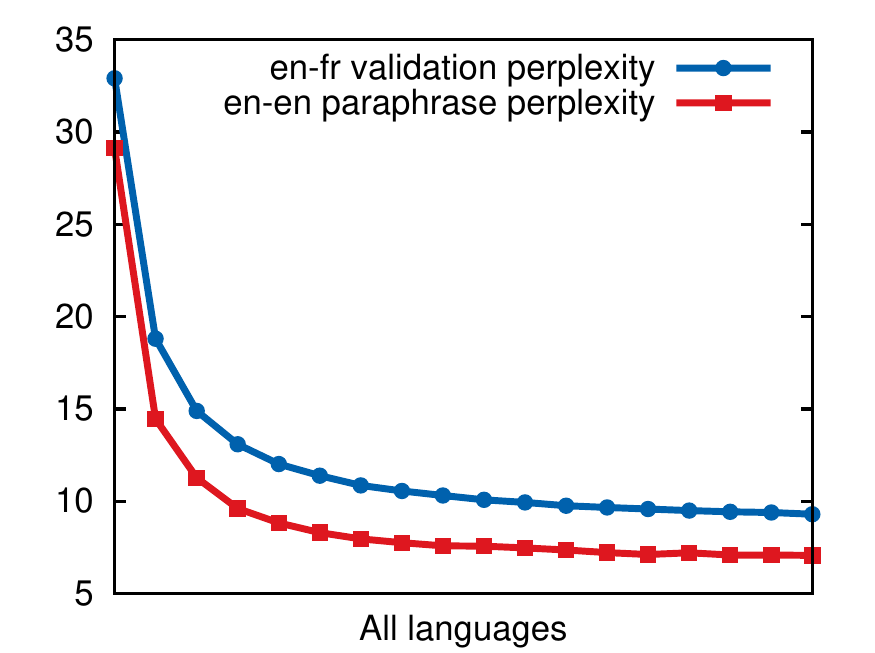}
\caption{Learning curves from three models (the bilingual English-French model, a trilingual model and a multilingual one): Perplexity on Bible data, English-French in validation (blue) and English paraphrases in testing (red). Note the different scales.}
\label{fig:learning-curves}
\end{figure*}

\subsection{Test data}

For our experiments, we apply test sets from two domains. One of them represents in-domain data from the Bible collection that covers 998 verses from the New Testament that we held out of training and development sets.  Our second test set comes from a very different source, namely data collected from user-contributed translations that are on-line in the Tatoeba database.\footnote{\url{https://tatoeba.org/eng/}} They include everyday expressions with translations in a large number of languages. As the collection includes translation alternatives, we can treat them as paraphrases of each other. We extracted altogether 3,873 pairs of synonymous sentences in English.

From both test data sources, we create a single-reference test set for paraphrase recognition and a multi-reference test set for paraphrase generation. The single-reference Bible test set uses the \textit{Standard} English Bible as the source, and the \textit{Common} English Bible\footnote{CEB is an ambitious new translation rather than a revision of other translations (https://www.biblegateway.com).}  as the reference.  The multi-reference Bible test set uses the \textit{Amplified} Bible as the source (the first one on our list), and all 18 other English Bibles as the references.

The Tatoeba single-reference test set contains all 3,873 synonymous sentence pairs. For the multi-reference test set, we filtered the data to exclude near-identical sentence pairs by expanding contractions (like "I'm" to "I am") that are quite common in the data and removed all pairs that differ only in punctuation after that procedure. Furthermore, we merged alternatives of the same sentence into synonym sets and created, thus, a multi-reference corpus for testing containing a total of 2,444 sentences with their references.

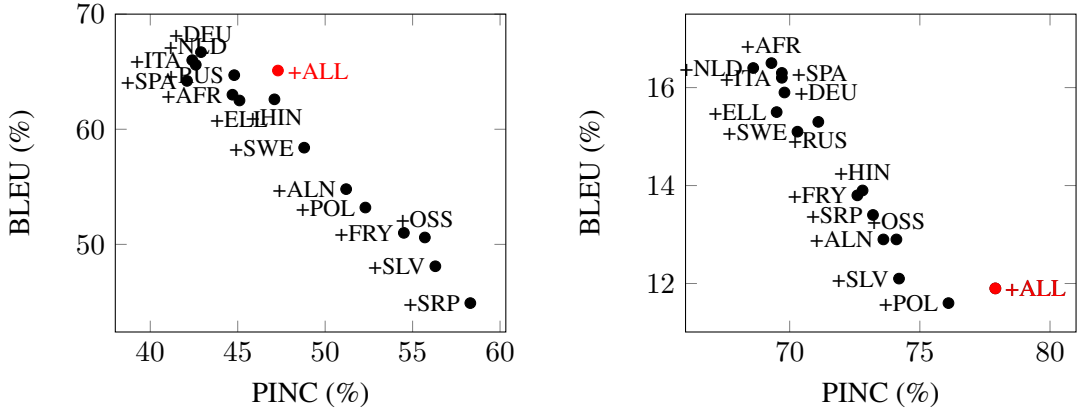
\begin{figure*}[ht]
\centering
\begin{tikzpicture}
\begin{axis}[ymax=70, xmin=38, xlabel=PINC (\%), ylabel=BLEU (\%), width=0.42\textwidth]
\addplot[scatter,only marks,mark=*,nodes near coords,visualization depends on={value \thisrow{anchor}\as\myanchor}, every node near coord/.append style={anchor=\myanchor,font=\small},point meta=explicit symbolic] table[meta=label] {
x y label anchor
44.7 63.0 +AFR east
51.2 54.8 +ALN east
42.9 66.7 +DEU south
45.1 62.5 +ELL north east
54.5 51.0 +FRY east
47.1 62.6 +HIN north east
42.4 66.0 +ITA east
42.6 65.6 +NLD south east
55.7 50.6 +OSS south east
52.3 53.2 +POL east
44.8 64.7 +RUS east
56.3 48.1 +SLV east
42.1 64.2 +SPA east
58.3 44.9 +SRP east
48.8 58.4 +SWE east
    };
\addplot[scatter,only marks,mark=*,nodes near coords,visualization depends on={value \thisrow{anchor}\as\myanchor}, every node near coord/.append style={anchor=\myanchor,font=\small,color=red},mark options={color=red},point meta=explicit symbolic] table[meta=label] {
x y label anchor
47.3 65.1 +ALL west
    };
\end{axis}
\end{tikzpicture}%
\hspace{5mm}
\begin{tikzpicture}
\begin{axis}[ymax=17.5, xmin=66, xmax=81, xlabel=PINC (\%), ylabel=BLEU (\%), width=0.42\textwidth]
\addplot[scatter,only marks,mark=*,nodes near coords,visualization depends on={value \thisrow{anchor}\as\myanchor}, every node near coord/.append style={anchor=\myanchor,font=\small},point meta=explicit symbolic] table[meta=label] {
x y label anchor
69.3 16.5 +AFR south
73.6 12.9 +ALN east
69.8 15.9 +DEU west
69.5 15.5 +ELL east
72.6 13.8 +FRY east
72.8 13.9 +HIN south
69.7 16.2 +ITA east
68.6 16.4 +NLD east
74.1 12.9 +OSS south
76.1 11.6 +POL east
71.1 15.3 +RUS north
74.2 12.1 +SLV east
69.7 16.3 +SPA west
73.2 13.4 +SRP east
70.3 15.1 +SWE east
77.9 11.9 +ALL west
    };
\addplot[scatter,only marks,mark=*,nodes near coords,visualization depends on={value \thisrow{anchor}\as\myanchor}, every node near coord/.append style={anchor=\myanchor,font=\small,color=red},mark options={color=red},point meta=explicit symbolic] table[meta=label] {
x y label anchor
77.9 11.9 +ALL west
    };
\end{axis}
\end{tikzpicture}

\caption{Paraphrase BLEU vs. PINC scores for the Bible test set (left) and the Tatoeba test set (right).}
\label{fig:bleu-pinc}
\end{figure*}

\section{Results}

We evaluate the models on two tasks: (1) paraphrase recognition and (2) paraphrase generation. The following sections summarize our main findings in relation to these two tasks. We also evaluated the actual translation performance to ensure that the models are properly trained. The results of that test are listed in the supplementary material.

\subsection{Paraphrase Recognition}
\label{sec:pararec}

First of all, we would like to know how well our translation models are capable of handling paraphrased sentences. For this, we compute perplexity scores of the various models when observing English output sentences for given English input coming from the two paraphrase test sets. The intuition is that models with a higher level of semantic abstraction in the encoder should be less surprised by seeing paraphrased sentences on the decoder side, which will result in a lower perplexity.

Let us first look at the in-domain data from our Bible test set. Figure~\ref{fig:ppl} (left half) illustrates the reduction in perplexity when adding languages to our bilingual model. The figure is sorted by decreasing perplexities. While the picture does not reveal any clear pattern about the languages that help the most, we can see that they all contribute to an improved perplexity in comparison to the bidirectional English-French model. Breton is clearly the least useful language, without doubt due to the size of that language in our collection.  Note that a further 5\% perplexity reduction over the best trilingual model is achieved by the model that combines all languages (perplexity of 7.23, which is very close to the lower bound of 6.05).

The picture is similar but with a slightly different pattern on out-of-domain data. Figure~\ref{fig:ppl} (right half) shows the same plot for the Tatoeba test set with languages sorted in the same order as in the previous figure. Adding languages helps again, which is re-assuring, but the amount is less pronounced and further away from the lower bound (which is, however, to be expected in this setup). Again, Breton is not helping as much.  Furthermore, in the out-of-domain case, the model combining all languages actually does not improve the perplexity any further (the value of 42.63 is similar to other trilingual models), which is most probably due to the strong domain mismatch that influences the scores significantly.

To further demonstrate the problems of the bilingual model to learn proper semantic representations that can be used for paraphrase detection, we can also have a look at the learning curves in Figure~\ref{fig:learning-curves}. The first plot nicely shows that the perplexity scores on paraphrase data do not follow the smooth line of the validation data in English and French whereas the models that include auxiliary languages have the capability to improve the model with respect to paraphrase recognition throughout the training procedure in a similar way as the main objective (translation) is optimized.  The model that combines all languages achieves by far the lowest paraphrase perplexity. Learning curves of other trilingual models look very similar to the one included here.

\begin{table*}
\begin{minipage}[t]{0.60\textwidth} 
\small
\begin{tabular}{|lp{0.82\linewidth}|}
\toprule
Source & But even as he was on the road going down, his servants met him and reported, saying, Your son lives! \\
+NLD &  And as he was on the road, his servants went down with him, and reported, saying, Thy son lives! \\
+SPA &  But as it was on the road, his servants came to him and told him, “Your own Son lives!” \\
+ALL & And while he was on the way, his servants came to him, saying, “Your son lives!” \\
\midrule
Source & Give attention to this! Behold, a sower went out to sow. \\
+AFR & Pay attention to this! Behold, the sower went out to sow. \\
+ALL & Take care of this. Behold, a sower went out to sow. \\
+BRE & Give attention to this! For, look! un semeur sortit pour semer. \\
+DEU & Listen to this! Behold, a sower went out to sow. \\
\bottomrule
\end{tabular}
\end{minipage}
\begin{minipage}[t]{0.33\textwidth} 
\small
\begin{tabular}{|lp{0.82\linewidth}|}
\toprule
Source & He slept soundly. \\
Eng-Fra & Et il se prosterna devant soi.\\
+BRE & And, behold, he rose up quickly.\\
+DEU & And he began to sleep.\\
+ELL & He was sleeping.\\
+ALL & And when he had died, he was asleep.\\
\midrule
Source & She has no brothers. \\
Eng-Fra & Elle n'a point de frères. \\
+BRE & Or, elle n'a pas de frères. \\
+DEU & For she has no brothers. \\
+OSS & No, brothers.\\
+ALL & You have no brothers. \\
\bottomrule
\end{tabular}
\end{minipage}
\caption{Examples of generated Bible (left) and Tatoeba (right) paraphrases.}
\label{fig:gen-examples}
\end{table*}

\subsection{Paraphrase Generation}

This second experiment aims at testing the capacity of the NMT models to generate paraphrases of the input instead of translations.  The hypothesis is that the generated sentences will preserve the meaning of the input, but not necessarily the same form, such that the generated sentences can be viewed as genuine paraphrases of the input sentences.

Good paraphrase models should produce sentences that are as close as possible to one of the references, yet as different as possible from the source. The first part can be measured by common machine translation metrics such as BLEU \citep{papineni-etal-acl02}, which supports multiple references. The second part can be measured by specific paraphrase quality metrics such as PINC \citep{chen-dolan-acl11}, which computes the proportion of non-overlapping n-grams between the source and the generated paraphrase. Good paraphrases should thus obtain high BLEU as well as high PINC scores on some paraphrase test set.

Figure~\ref{fig:bleu-pinc} plots BLEU scores against PINC scores for the two test sets (lowercased and ignoring punctuations), the alternative English translations in the heldout data from the Bible and the Tatoeba paraphrase set.  We exclude the bilingual model and the Breton model from the graphs, as they have BLEU scores close to 0 and PINC scores close to 100\% due to the output being generated in the wrong language.

The figures show a more or less linear correlation between BLEU and PINC. This is expected to a certain extent, as there is a clear trade-off between producing varied sentences (higher PINC) and preserving the meaning of the source sentence (higher BLEU). However, we find that the model containing all languages shows the overall best performance (e.g., according to the arithmetic mean of PINC and BLEU). This suggests that a highly multilingual model provides indeed more abstract internal representations that eventually lead to higher-quality paraphrases. We also conclude that additional languages with large and diverse (i.e., many different Bibles) datasets are better at preserving the meaning of the source sentence. However, there is no obvious language family or similarity effect.

The Tatoeba test set yields much lower BLEU scores than the Bible test set, due to the large number of unseen words and constructions, and also because the Tatoeba test set has only an average of 1.1 reference paraphrases per sentence, whereas the Bible test set has 18 references for each verse. This is most probably also the reason why the multilingual model including all languages ({\em ALL}) performs worse than most other models in terms of BLEU scores for the Tatoeba paraphrase test. It is highly likely that plausible paraphrases are not part of the test set if it only includes one or very few references like it is the case with Tatoeba, which is obviously a short-coming of BLEU as a metric for paraphrase evaluation.

Table~\ref{fig:gen-examples} shows some examples of paraphrases generated from the Bible and Tatoeba test set. One can see that different models tend to produce different paraphrases while preserving the general meaning of the source sentence at least in the case of the Bible data. Tatoeba is more problematic due to the domain mismatch and we will come back to that issue in the discussions further down.

One caveat is that paraphrase generation could trivially be achieved by copying the input to the output especially when evaluating the results using BLEU. Therefore, we also measured the percentage of identical copies that each model produces leaving out punctuations and lowercasing the data. The results show that copying is a rare case for the multilingual models and the input is only matched in at most 1.4\% of the cases (for Bible data) and at most 5.1\% of the cases in the Tatoeba test set.  However, adding English-English training data changes this behaviour dramatically, increasing the copying effect to over 70\% of the cases in both test sets, which breaks the use of such models as a paraphrase generator. This happens even though we train on pairs of different Bible translations into English, effectively training a paraphrase model with supervised learning. Details of this evaluation are given in the supplementary material.

Finally, we can also observe the effect of domain mismatch between the training data and the Tatoeba test set.  A considerable proportion of the test vocabulary refers to contemporary objects which obviously do not appear in the Bible training corpus, and it will, thus, be difficult for the model to generate adequate paraphrases.  A few examples of sentences containing out-of-vocabulary words are shown in Figure~\ref{fig:ex-tatoeba-para}. They indicate that the models are able to partially grasp the semantics of concepts and sentences often trying to replace unknown expressions with creative but reasonable alternatives coming from the context of the Bible. However, this observation calls for a more systematic evaluation of the semantic similarity of paraphrases than it is done by n-gram overlap with reference paraphrases, which is, unfortunately, out of the scope of this paper.

\begin{figure}
\small
\begin{tabular}{lp{0.75\linewidth}}
\toprule
Source & Have you never eaten a kiwi? \\
+AFR & Have you not eaten sour grapes? \\
\midrule
Source & Do you have a cellphone? \\
+HIN & Do you have a scorpion? \\
\midrule
Source & Do your children speak French? \\
+SPA & Do your children speak Greek? \\
\midrule
Source & Could I park my car here? \\
+ITA & Do I get up here with my cavalry? \\
\midrule
Source & Birds fly. \\
+DEU & The flying creatures shall fly away .\\
\bottomrule
\end{tabular}
\caption{Examples of generated Tatoeba paraphrases.}
\label{fig:ex-tatoeba-para}
\end{figure}

\section{Conclusions}

We have presented a study on the meaning representations that can be learned from multilingual data sets. We show that additional linguistic diversity lead to stronger abstractions and we verify our intuitions with a paraphrase scoring task that measures perplexity of multilingual sequence-to-sequence models. We also investigate the ability of translation models to generate paraphrases and conclude that this is indeed possible with promising results even without diversified decoders. In the future, we will try to push the model further to approach truly language-independent meaning representation based on massively parallel data sets as additional translational grounding. We will also study the model with bigger and less homogenous data sets and compare it to other approaches to paraphrase generation including pivot-based back-translation models. Furthermore, we will test sentence representations obtained by multilingual NMT models with additional downstream tasks to further support the main claims of the paper.

\section*{Acknowledgments}

The work in this paper is supported by the Academy of Finland through project 314062 from the ICT 2023 call on Computation, Machine Learning and Artificial Intelligence and the European Research Council (ERC) under the European Union’s Horizon 2020 research and innovation programme (grant agreement No 771113).  We would also like to acknowledge NVIDIA and their GPU grant.


\bibliographystyle{acl_natbib}
\bibliography{multipara}

\end{document}


\maketitle

\section{Translation} \label{sec:trans}

Table~\ref{tab:BLEU} summarizes the BLEU scores when testing on heldout data from the in-domain corpus (Bible) and the out-of-domain corpus (Tatoeba). For the former we used heldout data from the English standard Bible and the New-World Bible for French. For Tatoeba, we created a multi-reference test corpus from the English-French translations in the database that includes 1,068 English sentences with a total of 7,998 translations into French.

\begin{table}[h!]
\small
\centering
\begin{tabular}{|l|rr|rr|}
\hline
 \textbf{Training} & \multicolumn{2}{c|}{\textbf{Bible}} & \multicolumn{2}{c|}{\textbf{Tatoeba}} \\
\textbf{languages} & BLEU & $\Delta$ & BLEU & $\Delta$ \\
\hline
English--French & 21.29 &  & 15.62 & \\
\hline
+ Afrikaans  & 21.14 & -0.15 & 16.49 &  {\bf 0.87}\\
+ Albanian   & 21.22 & -0.07 & 15.82 &  {\bf 0.20}\\
+ Breton     & 20.91 & -0.38 & 15.43 & -0.19\\
+ German     & 20.77 & -0.52 & 14.63 & -0.99\\
+ Greek      & 20.87 & -0.42 & 15.43 & -0.19\\
+ Frisian    & 21.59 &  {\bf 0.30} & 15.52 & -0.10\\
+ Hindi      & 21.47 &  {\bf 0.18} & 15.07 & -0.55\\
+ Italian    & 21.40 &  {\bf 0.11} & 16.48 &  {\bf 0.86}\\
+ Dutch      & 21.18 & -0.11 & 16.30 &  {\bf 0.68}\\
+ Ossetian   & 20.84 & -0.45 & 17.11 &  {\bf 1.49}\\
+ Polish     & 21.05 & -0.24 & 17.18 &  {\bf 1.56}\\
+ Russian    & 21.00 & -0.29 & 15.49 & -0.13\\
+ Slovene    & 21.40 &  {\bf 0.11} & 16.30 &  {\bf 0.68}\\
+ Spanish    & 20.81 & -0.48 & 15.11 & -0.51\\
+ Serbian    & 21.44 &  {\bf 0.15} & 17.19 &  {\bf 1.57}\\
+ Swedish    & 20.64 & -0.65 & 16.85 &  {\bf 1.23}\\
\hline
{\bf average} & 21.11 & -0.18 & 16.03 & {\bf 0.41}\\
\hline
\end{tabular}
\caption{English to French translation quality in terms of BLEU scores, using the in-domain Bible test set (left half, single reference) and the out-of-domain Tatoeba test set (right half, multiple references). The columns marked with $\Delta$ show the absolute BLEU score difference compared to the baseline English--French model; improvements are highligted in bold face.}\label{tab:BLEU}
\end{table}

In-domain translation with multilingual models is on par with bilingual ones and ot-of-domain models show gains in the majority of the cases up to 1.57 BLEU points.

\section{Paraphrase Generation}

Table~\ref{tab:identical} lists the percentage of identical copies produced when using multilingual NMT models for paraphrase generation. In the comparison we discard punctuations and compare lowercased strings. Adding English-to-English paraphrased training data significantly increases the percentage.

\begin{table}[h!]
\small
\center
\begin{tabular}{|lrr|}
\hline
\textbf{Model} & \textbf{Bible} & \textbf{Tatoeba} \\
\hline
English--French & 0.0\% & 0.7\% \\
\hline
+ Afrikaans & 0.9\% & 4.8\% \\
+ Albanian & 0.7\% & 3.4\% \\
+ Breton & 0.0\% & 1.1\% \\
+ German & 1.4\% & 4.9\% \\
+ Greek & 1.1\% & 5.2\% \\
+ Frisian & 0.7\% & 4.3\% \\
+ Hindi & 0.9\% & 4.2\% \\
+ Italian & 1.2\% & 5.0\% \\
+ Dutch & 1.1\% & 5.1\% \\
+ Ossetian & 0.6\% & 3.5\% \\
+ Polish & 0.4\% & 2.8\% \\
+ Russian & 1.4\% & 4.7\% \\
+ Slovene & 0.6\% & 3.2\% \\
+ Spanish & 1.1\% & 5.5\% \\
+ Serbian & 0.5\% & 3.3\% \\
+ Swedish & 1.2\% & 4.9\% \\
\hline
+ All & 0.8\% & 2.0\% \\
\hline
+ English--English & 71.6\% & 70.0\% \\
\hline
\end{tabular}
\caption{Percentages of identical source and generated target sentences. Multilingual models produce significantly less copies of the input compared to the supervised paraphrase model trained on pairs of English Bible variants (last line).}
\label{tab:identical}
\end{table}